\documentclass{article}

\usepackage[final,nonatbib]{neurips_2019}
\usepackage[square,sort,comma,numbers]{natbib}

\usepackage[utf8]{inputenc} 
\usepackage[T1]{fontenc}    
\usepackage{hyperref}       
\usepackage{url}            
\usepackage{booktabs}       
\usepackage{amsfonts}       
\usepackage{nicefrac}       
\usepackage{microtype}      

\usepackage{caption}
\usepackage{graphicx}
\usepackage{algorithm}
\usepackage{algorithmic}
\usepackage{float}

\title{Faster Boosting with Smaller Memory}

\author{%
  Julaiti Alafate\\
  Department of Computer Science and Engineering\\
University of California, San Diego\\
La Jolla, CA 92093 \\
\texttt{jalafate@eng.ucsd.edu} \\
  \And
  Yoav Freund\\
  Department of Computer Science and Engineering\\
University of California, San Diego\\
La Jolla, CA 92093 \\
\texttt{yfreund@ucsd.edu} \\
}

\graphicspath{{figs/}}

\newtheorem{theorem}{Theorem}

\newcommand{\vx}{\vec{x}}
\newcommand{\Prob}[2]{P_{#1}\left[#2 \right]}
\newcommand{\Expect}[2]{E_{#1}\left[#2 \right]}
\newcommand{\cH}{\mathcal{H}}

\newcommand{\PhiEmp}{\widehat{\Phi}}
\newcommand{\corr}{\mbox{corr}}
\newcommand{\corrEmp}{\widehat{\corr}}
\newcommand{\err}{\mbox{err}}

\newcommand{\edge}{\gamma}
\newcommand{\edgeEmp}{\hat{\edge}}
\newcommand{\sign}{\mbox{sign}}

\newcommand{\Dist}{{\cal D}}

\newcommand{\n}{n}
\newcommand{\neff}{{n_{\mbox{\tiny eff}}}}

\newcommand{\Sparrow}{{\bf Sparrow}}
\newcommand{\XGBoost}{{\bf XGBoost}}
\newcommand{\LightGBM}{{\bf LightGBM}}

\newcommand{\weakRules}{{\cal W}}

\begin{document}

\maketitle

\begin{abstract} State-of-the-art implementations of boosting, such as
  XGBoost and LightGBM, can process large training sets extremely
  fast. However, this performance requires that the memory size is
  sufficient to hold a 2-3 multiple of the training set size.  This
  paper presents an alternative approach to implementing the boosted
  trees, which achieves a significant speedup over XGBoost and
  LightGBM, especially when the memory size is small. This is achieved
  using a combination of three techniques: early stopping, effective
  sample size, and stratified sampling. Our experiments demonstrate a
  10-100 speedup over XGBoost when the training data is too large to
  fit in memory.
\end{abstract}

\section{Introduction}\label{sec:intro}

Boosting~\cite{freund_decision-theoretic_1997,schapire_boosting:_2012},
and in particular gradient boosted trees~\cite{friedman_greedy_2001},
are some of the most popular learning algorithms used in practice.
There are several highly optimized implementations of boosting,
among which \XGBoost~\cite{chen_xgboost:_2016} and
\LightGBM~\cite{ke_lightgbm:_2017} are two of the most popular ones.
These implementations can train models with hundreds of trees using
millions of training examples in a matter of minutes.

However, a significant limitation of these methods is that all of the training examples are required to
store in the main memory. For LightGBM
this requirement is strict. XGBoost can operate in the disk-mode,
which makes it possible to use machines with smaller memory than the
training set size. However, it comes with a penalty in much longer
training time.

In this paper, we present a new implementation of boosted
trees\footnote{The source code of the implementation is released at
  \url{https://github.com/arapat/sparrow}.}.  This implementation can
run efficiently on machines whose memory sizes are much smaller than the
training set. It is achieved with no penalty in accuracy, and with a
speedup of 10-100 over XGBoost in disk mode.

Our method is based on on the observation that each boosting step
corresponds to an estimation of the gradients along the axis
defined by the weak rules. The common approach to performing
this estimation is to scan {\em all} of the training examples so as to
minimize the estimation error. This operation is very expensive especially when
the training set does not fit in memory.

We reduce the number of examples scanned in each boosting iteration by
combining two ideas. First, we use {\em early stopping}~\cite{wald_sequential_1973}
to minimize the number of examples
scanned at each boosting iteration. Second, we keep in memory only a
sample of the training set, and we replace the sample when the sample
in memory is a poor representation of the complete training set.
We exploit the fact that boosting tends to place large weights on a
small subset of the training set, thereby reducing the effectivity of
the memory-resident training set. We propose a measure for quantifying
the variation in weights called the {\em effective number of
  examples}. We also describe an efficient algorithm,  {\em
  stratified weighted sampling}.

Early stopping for Boosting was studied in previous
work~\cite{domingo_scaling_2000,bradley_filterboost:_2007}.  The other
two are, to the best of our knowledge, novel. In the following
paragraphs we give a high-level description of these three ideas,
which will be elaborated on in the rest of the paper.

\paragraph{Early Stopping} We use early stopping to reduce the
number of examples that the boosting algorithm reads from the memory
to the CPU. A boosting algorithm adds a weak rule to
the combined strong rule iteratively. In most implementations, the algorithm
searches for the {\em best} weak rule, which requires scanning {\em
  all} of the training examples. However, the theory of boosting
requires the added weak rule to be just {\em significantly better than
   random guessing}, which does not require
scanning {\em all} of the training examples. Instead, our approach is to
read just as many examples as needed to identify a weak rule that is
significantly better than random guessing.

Our approach is based on {\em sequential analysis} and {\em early
  stopping}~\cite{wald_sequential_1973}. Using sequential analysis
methods, we designed a stopping rule to decide when to stop reading more
examples without increasing the chance of over-fitting.
  
\paragraph{Effective Number of Examples}
Boosting assigns different weights
to different examples.
The weight of an example represents the magnitude of its ``influence''
on the estimate of the gradient.
However, when 
the weight distribution of a training set is dominated by a small number of
``heavy'' examples,
the variance of the gradient estimates is high.
It leads to over-fitting,
and effectively reduces the size of the training set.
We quantify
this reduction using the {\em effective number of examples},
$\neff$.
To get reliable estimates, $\neff$ should be close to the size
of the current training set in memory, $\n$.
When $\frac{\neff}{\n}$ is small,
we flush the current training set, and get a new sample
using {\em weighted sampling}.

\paragraph{Stratified Weighted Sampling} While there are well known methods
for weighted sampling, all of the existing methods (that we know of)
are inefficient when the weights are highly skewed.
In such cases, most of the scanned examples are rejected, which leads to
very slow sampling. To increase the sampling efficiency, we introduce
a technique we call {\em stratified weighted
  sampling}.
  It generates same sampled distribution while
guaranteeing that the fraction of rejected examples is no large than
$\frac{1}{2}$.

We implemented a new boosted tree algorithm with these three techniques, called \Sparrow.
We compared its performance to the
performance of \XGBoost\ and \LightGBM\ on two large datasets: one with 50 Million examples (the human
acceptor splice site dataset~\cite{sonnenburg_coffin_2010,
  agarwal_reliable_2014}), the other with over 600M examples (the bathymetry dataset~\cite{bathymetry}). We show that \Sparrow\ can achieve 10-20x speed-up over \LightGBM\ and \XGBoost\ especially in the limited memory settings.

The rest of the paper is organized as follows.
In Section~\ref{sec:related} we discuss
the related work.
In Section~\ref{sec:pre} we review the confidence-based boosting algorithm.
In Section~\ref{sec:theory} we describe the statistical theory behind the design of Sparrow.
In Section~\ref{sec:Algorithms} we describe
the design of our implementation. In Section~\ref{sec:experiments} we
describe our experiments. We conclude with the future work direction
in Section~\ref{sec:Conclusion}.

\section{Related Work}\label{sec:related}

There are several methods that uses sampling to reduce the training time
of boosting.
Both Friedman {\em et\,al.}~\cite{friedman_additive_2000} and
LightGBM~\cite{ke_lightgbm:_2017} use a fixed threshold to filter out the light-weight examples:
the former discards the examples whose weights are smaller than the threshold;
the later accepts all examples if their gradients exceed the threshold, otherwise
accepts them with a fixed probability.
Their major difference with Sparrow is that
their sampling methods are biased, while Sparrow does not change the original data distribution.
Appel {\em et\,al.}~\cite{appel2013quickly} uses small samples to prune weak rules
associated with unpromising features, and only scans all samples for evaluating the remaining ones.
Their major difference with Sparrow is that 
they focus on finding the “best” weak rule, while Sparrow tries 
to find a “statistically significant” one. Scanning over 
all example is required for the former, while using a 
stopping rule the algorithm often stops after reading a 
small fraction of examples.

The idea of accelerating boosting with stopping rules is also studied
by
Domingo and Watanabe~\cite{domingo_scaling_2000} and Bradley and
Schapire~\cite{bradley_filterboost:_2007}.
Our contribution is in using a
tighter stopping rule. Our stopping rule is tighter because it takes
into account the dependence on the variance of the sample weights.

There are several techniques that speeds up boosting by taking advantage of
the sparsity of the
dataset~\cite{chen_xgboost:_2016,ke_lightgbm:_2017}.
We will consider those techniques in future work.

\newcommand{\Var}[1]{\mathrm{Var} \left[ #1 \right]}

\section{Confidence-Rated Boosting}\label{sec:pre}
We start with a brief description of the confidence-rated boosting
algorithm under the AdaBoost framework (Algorithm 9.1 on the page 274 of
\cite{schapire_boosting:_2012}). 

Let $\vx \in X$ be the feature vectors and let the output be $y \in
Y= \{-1,+1\}$. For a joint distribution $\Dist$ over
$X \times Y$, our goal is to find a classifier $c: X \to Y$ with small
error:
$$\err_{\Dist}(c) \doteq \Prob{(\vx,y)\sim \Dist}{c(\vx) \neq y}.$$

We are given a set $\cH$ of base classifiers (weak rules) $h:X \to
[-1,+1]$. We want to generate a {\em score function},
which is a {\em weighted} sum of
$T$ rules from $\cH$:
\[
S_T(\vx) = \left( \sum_{t=1}^T \alpha_t h_t(\vx) \right).
\]
The term $\alpha_t$ is the weights by which each base classifiers contributes
to the final prediction, and is decided by the specific boosting paradigm.

Finally, we have the strong classifier as the sign of the score function: $H_T =
\sign(S_T)$.  

AdaBoost can be viewed as a coordinate-wise gradient descent
algorithm~\cite{mason_boosting_1999}. The algorithm iteratively finds
the direction (weak rule) which maximizes the decrease of the average
potential function, and then adds this weak rule to $S_T$ with a
weight that is proportional to the magnitude of the gradient. The
potential function used in AdaBoost is $\Phi(\vx,y) = e^{-S_t(\vx)y}$. Other
potential functions have been studied
(e.g.~\cite{friedman_greedy_2001}).  In this work we focus on the
potential function used in AdaBoost.

We distinguish between two types of average potentials: the
expected potential or true potential:
\[
\Phi(S_t) = \Expect{(\vx,y) \sim \Dist}{e^{-S_t(\vx)y}},
\]
and the average potential or empirical potential:
\[
\PhiEmp(S_t) = \frac{1}{n} \sum_{i=1}^n e^{-S_t(\vx_i)y_i}.
\]

The ultimate goal of the boosting algorithm is to minimize the
expected potential, which determines the true error rate. However,
most boosting algorithms, including \XGBoost\ and \LightGBM, focus on
minimizing the empirical potential $\PhiEmp(S_t)$,
and rely on the
limited capacity of the weak rules to guarantee that the true
potential is also small. \Sparrow\ takes a different approach. It uses an estimator
of the true edge (explained below) to identify the weak rules that reduce the {\em
  true} potential with high probability.

Consider adding a weak rule $h_{t}$ to the score function $S_{t-1}$, we get 
$S_{t}=S_{t-1}+\alpha_{t} h_{t}$. Taking the partial derivative of the average potential
$\Phi$ with respect
to $\alpha_{t}$ we get
\begin{equation} \label{eqn:weights}
\left. \frac{\partial}{\partial \alpha_{t}}\right|_{\alpha_{t}=0} \Phi(S_{t-1}+\alpha_{t} h)
= \Expect{(\vx,y) \sim \Dist_{t-1}}{h(\vx) y}
\end{equation}
where
\begin{equation}~\label{eqn:Dt}
\Dist_{t-1} = \frac{\Dist}{Z_{t-1}} \exp\left( -S_{t-1}(\vx)y \right),
\end{equation}
and $Z_{t-1}$ is a normalization factor that makes $\Dist_{t-1}$ a
distribution.

Boosting algorithms performs coordinate-wise gradient descent on the
average potential where each coordinate corresponds to one weak rule. Using
equation (\ref{eqn:weights}), we can express the gradient with respect to
the weak rule $h$ as a correlation, which we call the {\em true edge}: 
\begin{equation} \label{eqn:true-edge}
\edge_{t}(h) \doteq \corr_{\Dist_{t-1}}(h) \doteq \Expect{(\vx,y) \sim \Dist_{t-1}}{h(\vx) y},
\end{equation}
which is not directly measurable. Given $n$ i.i.d.\ samples,
an unbiased estimate for
the true edge is the {\em empirical edge}:
\begin{equation} \label{eqn:emp-edge}
\edgeEmp_{t}(h) \doteq \corrEmp_{\Dist_{t-1}}(h)
\doteq 
\sum_{i=1}^n \frac{w_i}{Z_{t-1}} h(\vx_i) y_i,
\end{equation}
where  $w_i = e^{-S_{t-1}(\vx_i)}$ and $Z_{t-1} = \sum_{i=1}^n w_i$.

\section{Theory}\label{sec:theory}

To decrease the expected potential,
we want to find a weak rule with a large edge (and add it to the score function).
XGBoost and LightGBM do this by searching for
the weak rule
with the largest {\em empirical edge}.
 Sparrow finds a weak rule which,
with high probability, has a significantly large {\em true edge}.
Next, we explain
the statistical techniques for identifying such weak
rules while minimizing the number of
examples needed to compute the estimates.

\subsection{Effective Number of Examples}
\label{sec:effectiveSampleSize}
Equation~\ref{eqn:emp-edge} defines $\edgeEmp(h)$, which is an
unbiased estimate of $\edge(h)$. How accurate is this estimate?

A standard quantifier is the variance of the estimator.
Suppose the true edge of a weak rule $h$ is $\gamma$. Then the expected (normalized) correlation between the predictions of $h$ and the true labels,
$\frac{w}{Z} y h(x)$, is $2 \gamma$. The variance of this
correlation can be written as
$\frac{1}{n^2} \frac{ E {\left( w^2 \right)} }{E^2(w)} - 4\gamma^2$.
Ignoring the second term (because $\gamma$ is usually close to
zero) and the variance in $E(w)$, we approximate the
variance in the edge to be
\begin{equation} \label{eqn:variance}
 \mbox{Var}(\edgeEmp) \approx \frac{\sum_{i=1}^n w_i^2}{\left(\sum_{i=1}^n w_i\right)^2}.
\end{equation}

If all of the weights are equal then $\mbox{Var}(\edgeEmp) = 1/n$.
It corresponds to a standard deviation of $1/\sqrt{n}$ which is the
expected relation between the sample size and the error.

If the weights are not equal then the variance is larger and thus the
estimate is less accurate. We define the {\em effective number of examples}
$\neff$ to be $1/\mbox{Var}(\edgeEmp)$, specifically,
\begin{equation} \label{eqn:neff}
  \neff \doteq \frac{\left(\sum_{i=1}^n w_i\right)^2}{\sum_{i=1}^n w_i^2}.
\end{equation}

To see that the name ``effective number of examples'' makes sense, consider
$n$ weights where $w_1=\cdots=w_k=1/k$ and
$w_{k+1}=\cdots=w_{n}=0$. It is easy to verify that in this case
$\neff=k$ which agrees with our intuition, namely the examples with zero
weights have no effect on the estimate.

Suppose the memory is only large enough to store $\n$ examples. If $\neff
\ll \n$ then we are wasting valuable memory space on examples with
small weights, which can significantly increase the chance of over-fitting.
We can fix
this problem by using weighted sampling. In this way we repopulate memory with
$\n$ equally weighted examples, and make it possible to learn without
over-fitting.

\subsection{Weighted Sampling}

When \Sparrow\ detects that $\neff$ is much smaller than the memory
size $\n$, it clears the memory and collects a new sample from disk using
weighted sampling.

The specific sampling algorithm that \Sparrow\ uses is the {\em minimal
  variance weighted sampling}~\cite{kitagawa_monte_1996}. This method reads from disk one
example $(\vx,y)$ at a time, calculates the weight for that example
$w_i$, and accepts the example with the probability proportional to its weight.
Accepted examples are stored in  memory with the initial weights of
$1$. This increases the effective sample size from $\neff$ back to $\n$, thereby reduces the chance of over-fitting.

To gain some intuition regarding this effect, consider the following
setup of an imbalanced classification problem. Suppose that the training set size is $N=100,000$, of which $0.01 N$ are positive and $0.99 N$ are negative.
Suppose we can store $\n=2,000$ examples in memory. The number of the
memory-resident examples is $0.01 \n = 20$. Clearly, with such a small
number of positive examples, there is a danger of over-fitting. However,
an (almost) all negative rule is 99\% correct. If we then reweigh the examples
using the AdaBoost rule, we will give half of the total weight to the positives
and the other half to the negatives. The value of $\neff$ will drop
to about $80$. This would trigger a resampling step, which generates a
training set with 1000 positives and 1000 negatives. It
allows us to
find additional weak rules with little danger of over-fitting.

This process continues as long as \Sparrow\ is making progress and the
weights are becoming increasingly skewed. When the skew is large,
$\neff$ is small and \Sparrow\ resamples a new sample with uniform
weights.

Sparrow uses weighted sampling to achieve high disk-to-memory efficiency.
In addition, Sparrow also achieves high memory-to-CPU efficiency
by reading from memory
the minimal number of examples necessary to establish
that a particular weak rule has a significant edge.
This is done using
{\em sequential analysis} and {\em early stopping}.

\subsection{Sequential Analysis}\label{sec:seq-analysis}

Sequential analysis was introduced by
Wald in the 1940s~\cite{wald_sequential_1973}.
Suppose we want to estimate the expected loss of a
model. In the standard large deviation analysis, we assume that the
loss is bounded in some range, say $[-M,+M]$, and that the size of the
training set is $n$. This implies that the standard deviation of the
training loss is at most $M/\sqrt{n}$. In order to make this standard
deviation smaller than some $\epsilon>0$, we need that $n >
(M/\epsilon)^2$. While this analysis is optimal in the worst case, it
can be improved if we have additional information about the standard
deviation. We can glean such information from the observed losses by
using the following sequential analysis method.

Instead of choosing
$n$ ahead of time, the algorithm computes the loss one example at a
time. It uses a {\em stopping rule} to decide whether, conditioned on
the sequence of losses seen so far, the difference between the average
loss and the true loss is smaller than $\epsilon$ with large probability.
The result is that when the standard
deviation is significantly smaller than $M/\sqrt{n}$,
the number of examples
needed in the estimate is much smaller than
$(M/\epsilon)^2$.

We uses a stopping rule based on Theorem 1
in Appendix~\ref{sec:balsubramani},
which depends on both the mean and the variance of the
weighted correlation~\cite{balsubramani_sharp_2014}.
Fixing the current strong rule $H$ (i.e.\ the score function), we define a
(unnormalized) weight for each example, denoted as $w(\vx,y) = e^{-H(x)y}$.
Consider a
particular candidate weak rule $h$ and a sequence of labeled examples
$\{ (\vx_1,y_1),(\vx_2,y_2),\ldots \}$.
For some $\gamma > 0$,
we define two cumulative quantities (after seeing $n$ examples from the sequence):
\begin{equation} \label{eqn:stopping-rule-statistics}
  M_t \doteq \sum_{i=1}^n w(\vx_i,y_i) (h_t(\vx_i) y_i - \gamma), \mbox{ and }
  V_t \doteq  \sum_{i=1}^n w(\vx_i,y_i)^2.
\end{equation}
$M_t$ is an estimate of the difference between
the true correlation of $h$ and $\gamma$.
$V_t$ quantifies the variance of this estimate.

The goal of the stopping rule is
to identify a weak rule $h$ whose true edge is larger than $\gamma$.
It is defined to be $t > t_0$ and
\begin{equation} \label{eqn:stopping_rule}
M_t > C \sqrt{V_t(\log\log {V_t \over M_t}+ B)},
\end{equation}
where $t_0, C$, and $B$ are parameters.
If both conditions of the stopping rule are true, we claim
that the true edge of $h$ is larger than $\gamma$ with high probability.
The proof of this test can be found in~\cite{balsubramani_sharp_2014}.

Note that our stopping rule is correlated with
the cumulative variance $V_t$,
which is basically the same as $1/\neff$. If $\neff$ is large,
say $\neff = \n$ when a new sample is placed in memory,
the stopping rule stops quickly.
On the other hand, when the weights diverge,
$\neff$ becomes smaller than $\n$, and the stopping
rule requires proportionally more examples before stopping.

The relationship between martingales, sequential analysis, and stopping rules has been  studied in previous work~\cite{wald_sequential_1973}.
Briefly, when the edge of a rule is smaller than $\gamma$, then the sequence is a supermartingale. If it is larger than $\gamma$, then it is a submartingale. { The only assumption is that the examples are sampled i.i.d.}.
Theorem~\ref{thm:balsubramani} in Appendix~\ref{sec:balsubramani} guarantees two things about the stopping rule defined in Equation~\ref{eqn:stopping_rule}: (1) if the true edge is smaller than $\gamma$, the stopping rule will never fire (with high probability); (2) if the stopping rule fires, the true edge of the rule $h$ is larger than $\gamma$.

\section{System Design and Algorithms} \label{sec:Algorithms}

\begin{figure}
\centering
    \includegraphics[width=1.0\textwidth]{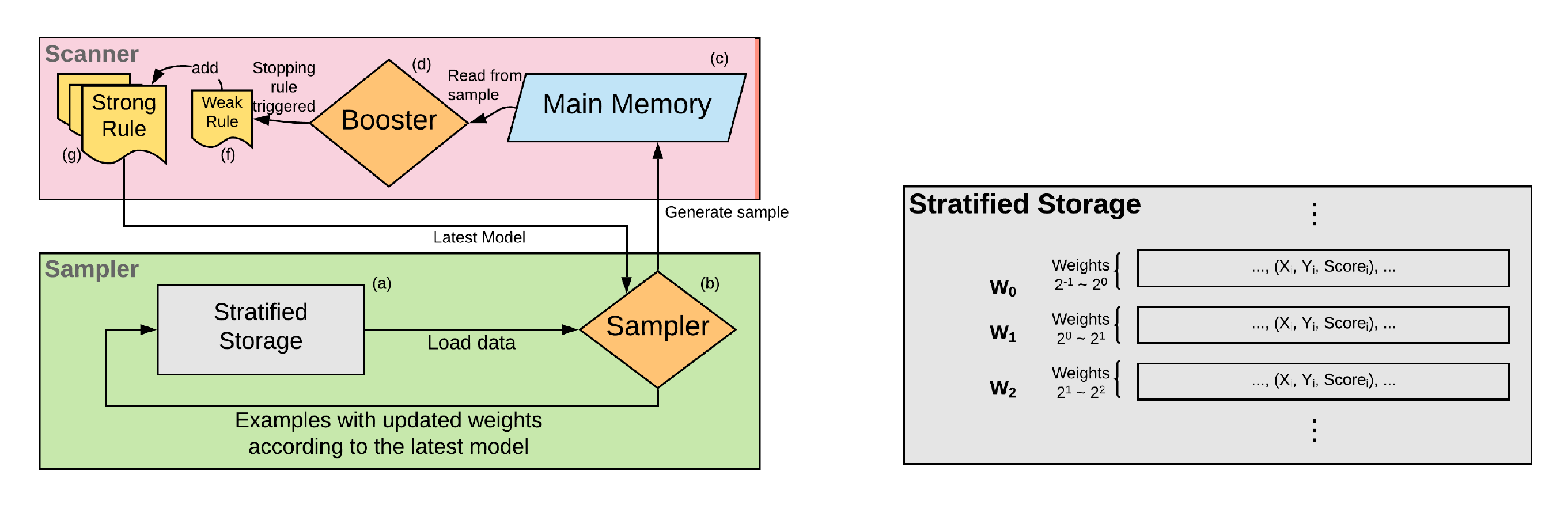}
    \caption{The \Sparrow\ system architecture.
    Left: The workflow of the Scanner and the Sampler. Right: Partitioning of the examples stored in disk according to their weights.}\label{fig:architecture}
    \vspace{0pt}
\end{figure}

In this section we describe \Sparrow. As Sparrow consists of a number
of concurrent threads and many queues, we chose to implement it using
the {\bf Rust} programming language for the benefits of its memory-safety and thread-safety guarantees~\cite{klabnik2018rust}.

We use a bold letter in parenthesis to refer the corresponding
component in the workflow diagram in Figure~\ref{fig:architecture}.
We also provide the pseudo-code in the Appendix~\ref{appendix:pseudocode}.

The main procedure of \Sparrow\ generates a sequence
of weak rules $h_1,\ldots,h_k$ and combines them into a strong rule
$H_k$. 
It calls two subroutines that execute in parallel:
a {\bf Scanner} and a {\bf Sampler}.

\paragraph*{Scanner}
The task of a scanner (the upper part of the workflow diagram in Figure~\ref{fig:architecture})
is to read training examples sequentially and stop
when it has identified one of the rules to be a {\em good} rule.

At any point of time, the scanner maintains the current strong
rule $H_t$, a set of candidate weak rules $\weakRules$, and a target
edge $\gamma_{t+1}$.
For example, when training boosted decision trees,
the scanner maintains the current strong rule $H_t$ which consists of a set of
decision trees, a set of candidate weak rules $\weakRules$
which is the set of candidate splits on all features, and $\gamma_{t+1} \in (0.0, 0.5)$. 

Inside the scanner, a booster {\bf (d)} scans the training examples stored in
main memory {\bf (c)} sequentially, one at a time. It computes the weight of the
read examples using $H_t$ and then updates a running estimate of the edge
of each weak rule $h \in \weakRules$ accordingly.
Periodically, it feeds these running estimates into the stopping rule,
and stop the scanning when the stopping rule fires.

The stopping rule is designed such that if it fires for $h_t$, then the true edge
of a particular weak rule $\gamma(h_{t+1})$ is, with high probability,
larger than the set threshold $\gamma_{t+1}$. The booster then adds the
identified weak rule $h_{t+1}$ {\bf (f)} to the current strong rule $H_t$
to create a new strong rule $H_{t+1}$ {\bf (g)}.
The booster decides the weight of the weak rule $h_{t+1}$ in $H_{t+1}$ based on $\gamma_{t+1}$ (lower bound of its accuracy).
It could underestimate the weight. However,
if the underestimate is large, the weak rule $h_{t+1}$ is likely to
be ``re-discovered'' later which will effectively increase its weight.

Lastly, the scanner falls into the \textit{Failed} state if after exhausting
all examples in the current sample set, no weak rule with an advantage
larger than the target threshold $\gamma_{t+1}$ is detected.  When it happens,
the scanner shrinks the value of $\gamma_{t+1}$
and restart scanning.  More precisely, it keeps track of the
empirical edges $\edgeEmp{(h)}$ of all weak rules $h$.  When the
failure state happens, it resets the threshold $\gamma_{t+1}$ to just
below the value of the current maximum empirical edge of all weak
rules.

\begin{figure}
\begin{minipage}{.48\textwidth}
\centering
    \includegraphics[width=1.0\textwidth]{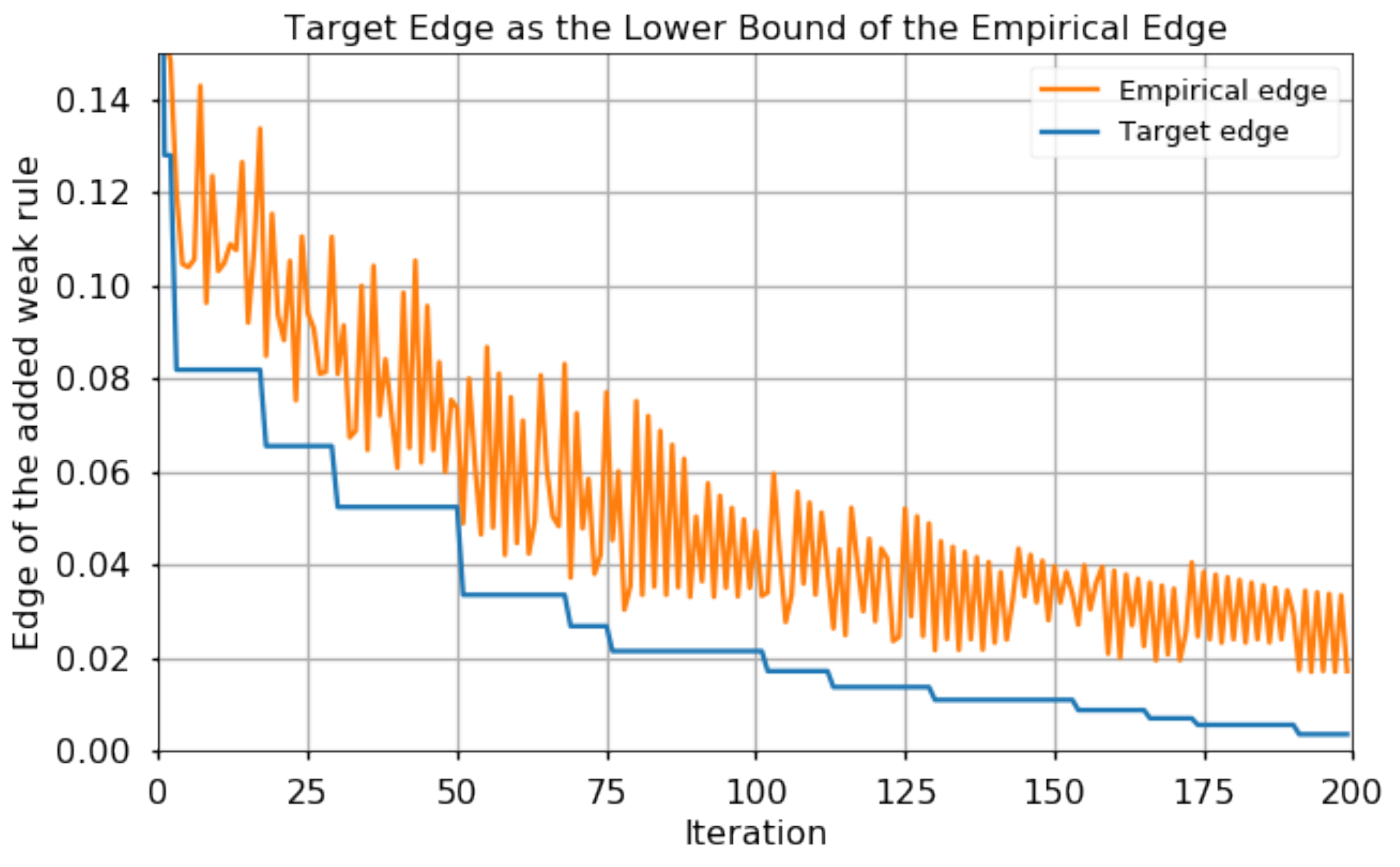}
    \caption{The empirical edge and the corresponding target edge $\gamma$
    of the weak rules being added to the ensemble.
    Sparrow adds new weak rules with a weight calculated
    using the value of $\gamma$ at the time of their
    detection, and shrinks $\gamma$ when it cannot detect a rule with an edge
    over $\gamma$.\label{fig:edge}}
    \vspace{8pt}
\end{minipage}
\hfill
\begin{minipage}{.48\textwidth}
\centering
    \includegraphics[width=1.0\textwidth]{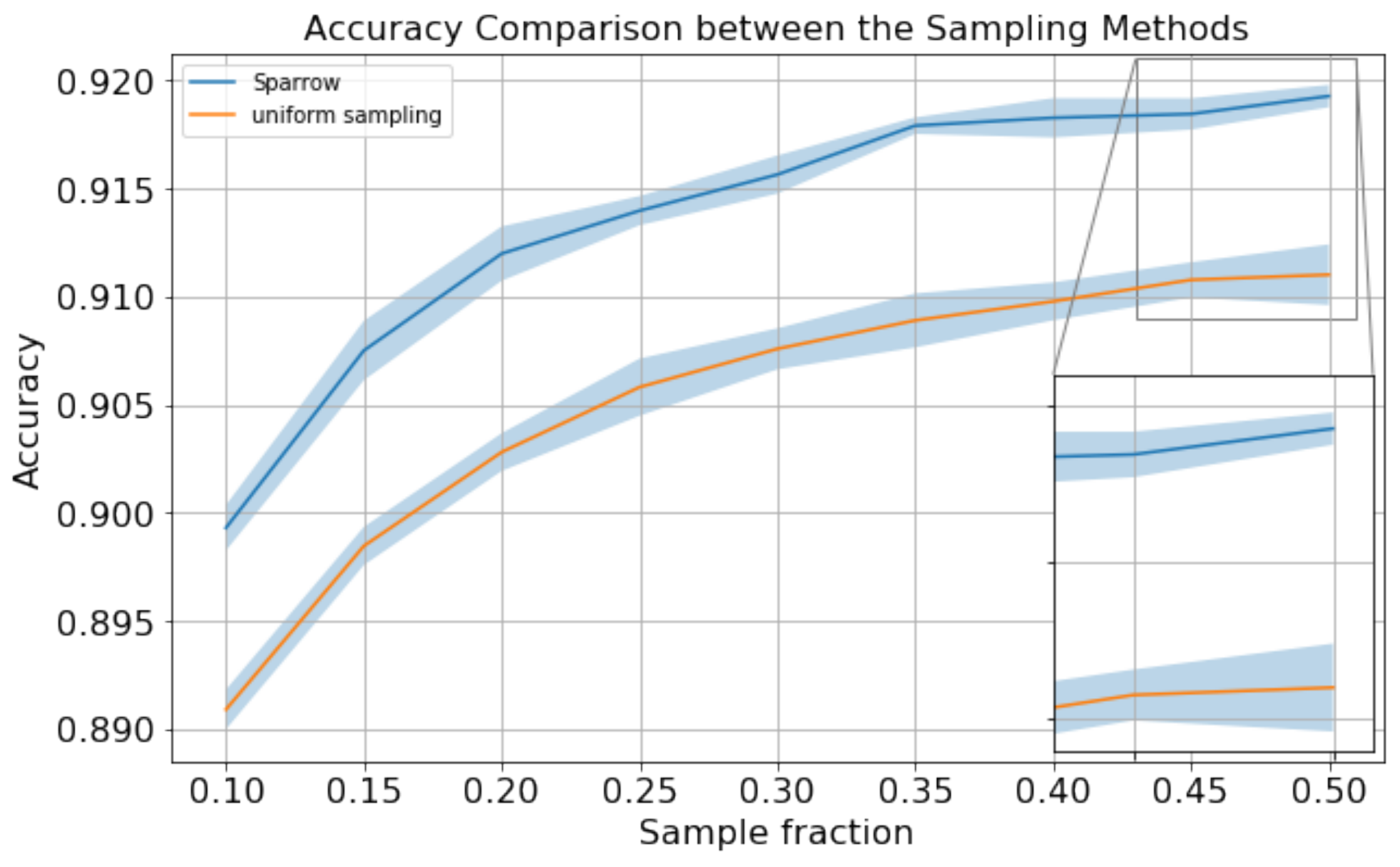}
    \caption{Accuracy comparison on the CoverType dataset.
	For uniform sampling, we trained XGBoost on a uniformly sampled dataset with the
	same sample fraction set in Sparrow. The accuracy is evaluated with same number
	of boosting iterations.}~\label{fig:covtype}
\end{minipage}
\end{figure}

To illustrate the relationship between the target threshold and the empirical edge of the
detected weak rule,
we compare their values in Figure~\ref{fig:edge}.
The empirical edge $\edgeEmp{(h_{t+1})}$ of the detected weak rules are usually larger than
$\gamma_{t+1}$. The weak rules are then added to the strong rule with
a weight corresponding to $\gamma_{t+1}$ (the lower bound for their true edges) to avoid over-estimation. Lastly, the value of $\gamma_{t+1}$ shrinks over
time when there is no weak rule with the larger edge exists.

\paragraph*{Sampler}
Our assumption is that the entire training dataset does
not fit into the main memory and is therefore stored in an external storage
{\bf (a)}. As boosting progresses, the weights of the examples become
increasingly skewed, making the dataset in memory effectively smaller.
To counteract that skew, { Sampler} prepares a {\em new}
training set, in which all of the examples have equal weights, by using
selective sampling. When the effective sample size $\neff$ associated
with the old training set becomes too small, the scanner stops using
the old training set and starts using the new one\footnote{The
  sampler and scanner can run in parallel on a multi-core machine, or run on two different machines. In our experiments, we keep them in one machine.}.

The sampler uses selective sampling by which we mean that the
probability of an example $(x,y)$ being added to the sample is
proportional to its weight $w(x,y)$. Each added example is assigned an initial
{ weight} of $1$.
{There are several known algorithms
  for selective sampling. The best known one is rejection sampling
  in which a biased coin is flipped for each example. We use a method
  known as \textit{minimal variance sampling}~\cite{kitagawa_monte_1996}
  because it produces less variation in the sampled set.}

\newcommand{\wmax}{w_{\mbox{\tiny max}}}
\newcommand{\wmean}{w_{\mbox{\tiny mean}}}

\paragraph*{Stratified Storage and Stratified Sampling}
The standard approach to sampling reads examples one at a time,
calculates the weight of the example, and accepts the
example into the memory with the probability proportional to its weight,
otherwise rejects the example. Let the largest weight be $\wmax$
and the average weight be $\wmean$, then the maximal rate at which
examples are accepted is $\wmean/\wmax$. If the weights are highly
skewed, then this ratio can be arbitrarily small, which means that
only a small fraction of the evaluated examples are then accepted. As
evaluation is time consuming, this process becomes a computation
bottleneck.

We proposed a stratified-based sampling mechanism to address this
issue (the right part of Figure~\ref{fig:architecture}).  It applies incremental
update to reduce the computational cost of making prediction with a
large model, and uses a stratified data organization to reduce the
rejection rate.

To implement incremental update we store for each example, whether it
is on disk or in memory, the result of the latest
update. Specifically, we store each training example
in a tuple $(x, y, H_l, w_l)$, where $x,y$ are the feature vector
and the label, $H_l$ is the last strong rule used to calculate the
weight of the example, and $w_l$ is the weight last calculated.
In this
way both the scanner and sampler only calculate over the incremental
changes to the model since the last time it was used to predict examples.

To reduce the rejection rate, we want the sampler to avoid reading examples
that it will likely to reject.
We organize examples in a stratified structure, where
the stratum $k$ contains examples whose weights are in $[2^k,2^{k+1})$.
This limits the skew of the weights of the examples
in each stratum so that $\wmean/\wmax \leq
\frac{1}{2}$.
In addition,
the sampler also maintains the (estimated) total weight of
the examples in each strata.
It then associates a probability with each stratum by
normalizing the total weights to $1$.

To sample a new example, the sampler first samples the next stratum to read,
then reads examples from the selected stratum until
one of them is accepted.
For each example, the sampler first updates its weight,
then decides whether or not to accept this example,
finally writes it back to the stratum it
belongs to according to its updated weight.
As a result, the reject rate is at
most $1/2$, which greatly improves the speed of sampling.

Lastly, since the stratified structure contains all of the examples, it is
managed mostly on disk, with a small in-memory buffer to
speed up I/O operations.

\section{Experiments}\label{sec:experiments}

In this section we describe the experiment results of
\Sparrow.
In all experiments, we use trees as weak rules.
First we use the forest cover type dataset~\cite{gama2003accurate}
to evaluate the sampling effectiveness.
It contains 581\,K samples.
We performed a 80/20 random split for training and testing.

In addition, 
we use two large datasets to evaluate the overall performance of Sparrow on large datasets.
The first large dataset is the splice site dataset for
detecting human acceptor splice site~\cite{sonnenburg_coffin_2010, agarwal_reliable_2014}.
We use the same training dataset of 50\,M samples as in the other work,
and validate the model on the testing data set of 4.6\,M samples.
The training dataset on disk takes over 39\,GB in size.
The second large dataset is the bathymetry dataset for
detecting the human mislabeling
in the bathymetry data~\cite{bathymetry}.
We use a training dataset of 623M samples, and validate the model
on the testing dataset of 83M samples. The training dataset takes 100\,GB on disk.
Both learning tasks are binary classification.

The experiments on large datasets are all conducted on EC2 instances with attached SSD storages from Amazon Web Services.
We ran the evaluations on five different instance types with increasing memory capacities, ranging from 8\,GB to 244\,GB (for details see Appendix~\ref{appendix:experiments}).

\subsection{Effectiveness of Weighted Sampling}

We evaluate the effectiveness of weighted sampling by comparing it to uniform sampling.
The comparison is over the model accuracy on the testing data
when both trained for 500 boosting iteration on
the cover type dataset. For both methods, we generate trees with depth 5 as weak rules.
In uniform sampling, we first randomly sample from the training data with
each sampling ratio, and use XGBoost to train the models.
We evaluated the model performance on the sample ratios ranging from $0.1$ to $0.5$, and
repeated each evaluation for 10 times.
The results are showed in Figure~\ref{fig:covtype}.
We can see that the accuracy of Sparrow is higher with the same number of boosting
iteration and same sampling ratio. In addition, the variance of the model accuracy is also
smaller. It demonstrates that the weighted sampling method used in Sparrow is more effective and
more stable than uniform sampling.

\subsection{Training on Large Datasets}

We compare Sparrow on the two large datasets, and use XGBoost and LightGBM for the baselines
since they out-perform other boosting implementations~\cite{chen_xgboost:_2016,ke_lightgbm:_2017}.
The comparison was done in
terms of the reduction in the exponential loss, which is what boosting
minimizes directly, and in terms of AUROC, which is often more
relevant for practice.
We include the data loading time in the reported training time.

There are two popular tree-growth algorithms: depth-wise and leaf-wise~\cite{shi2007best}.
Both \Sparrow\ and LightGBM grow trees leaf-wise.
XGBoost uses the depth-wise method by default.
In all experiments, we grow trees with at most $4$ leaves, or depth two.
We choose to train smaller trees in these experiments
 since the training take very long time otherwise.

For XGBoost, we chose approximate greedy algorithm which is its fastest training method.
LightGBM supports using sampling in the training,
which they called \textit{Gradient-based One-Side Sampling} (GOSS).
GOSS keeps a fixed percentage of examples with large gradients,
and randomly sample from remaining examples.
We selected GOSS as the tree construction algorithm for LightGBM.
In addition, we also enabled the \texttt{two\_round\_loading} option in LightGBM to 
reduce  its memory footprint.

Both XGBoost and LightGBM take advantage of the data sparsity for further speed-up training.
Sparrow does not deploy such optimizations in its current version,
which puts it in disadvantage.

The memory requirement of \Sparrow\ is decided by the sample size,
which is a configurable parameter.
XGBoost supports external memory training when the memory is too small to fit the
training dataset.
The in-memory version of XGBoost is used for training whenever possible.
If it runs out of memory,
we trained the model using the external memory version of XGBoost instead.
Unlike XGBoost, LightGBM does not support external memory execution.

Lastly, 
all algorithms in this comparison optimize the exponential loss as defined in AdaBoost.

\begin{figure}
\begin{minipage}{.48\textwidth}
	{
    \centering
    \includegraphics[width=1.0\textwidth]{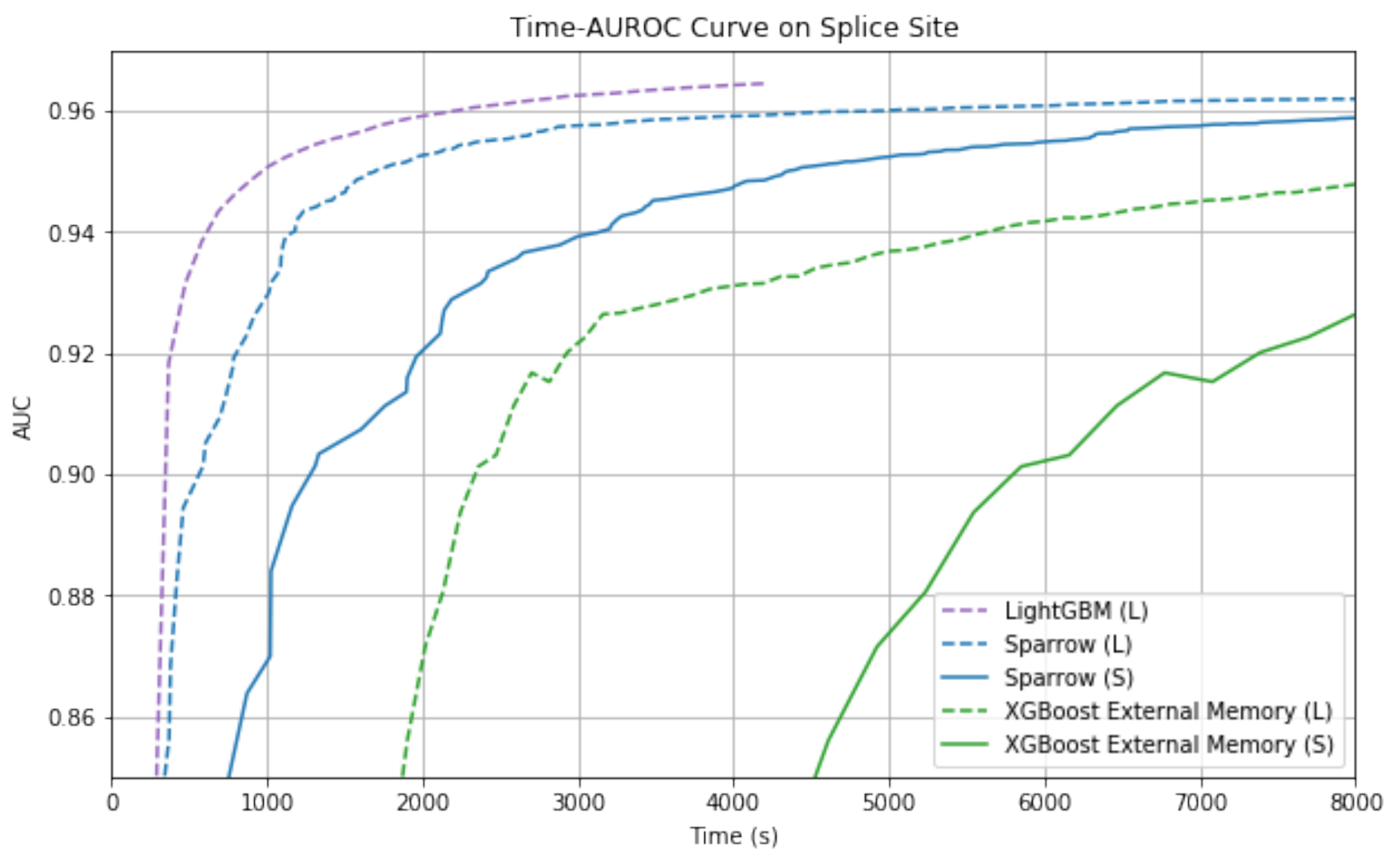}
    }
    \captionof{figure}{Time-AUROC curve on the splice site detection dataset, higher is better, clipped on right and bottom
    The (S) suffix is for training on 30.5\,GB memory, and the (L) suffix is for training on 61\,GB memory.}~\label{fig:splice-auroc}
\end{minipage}
\hfill
\begin{minipage}{.48\textwidth}
	{
    \centering
    \includegraphics[width=1.0\textwidth]{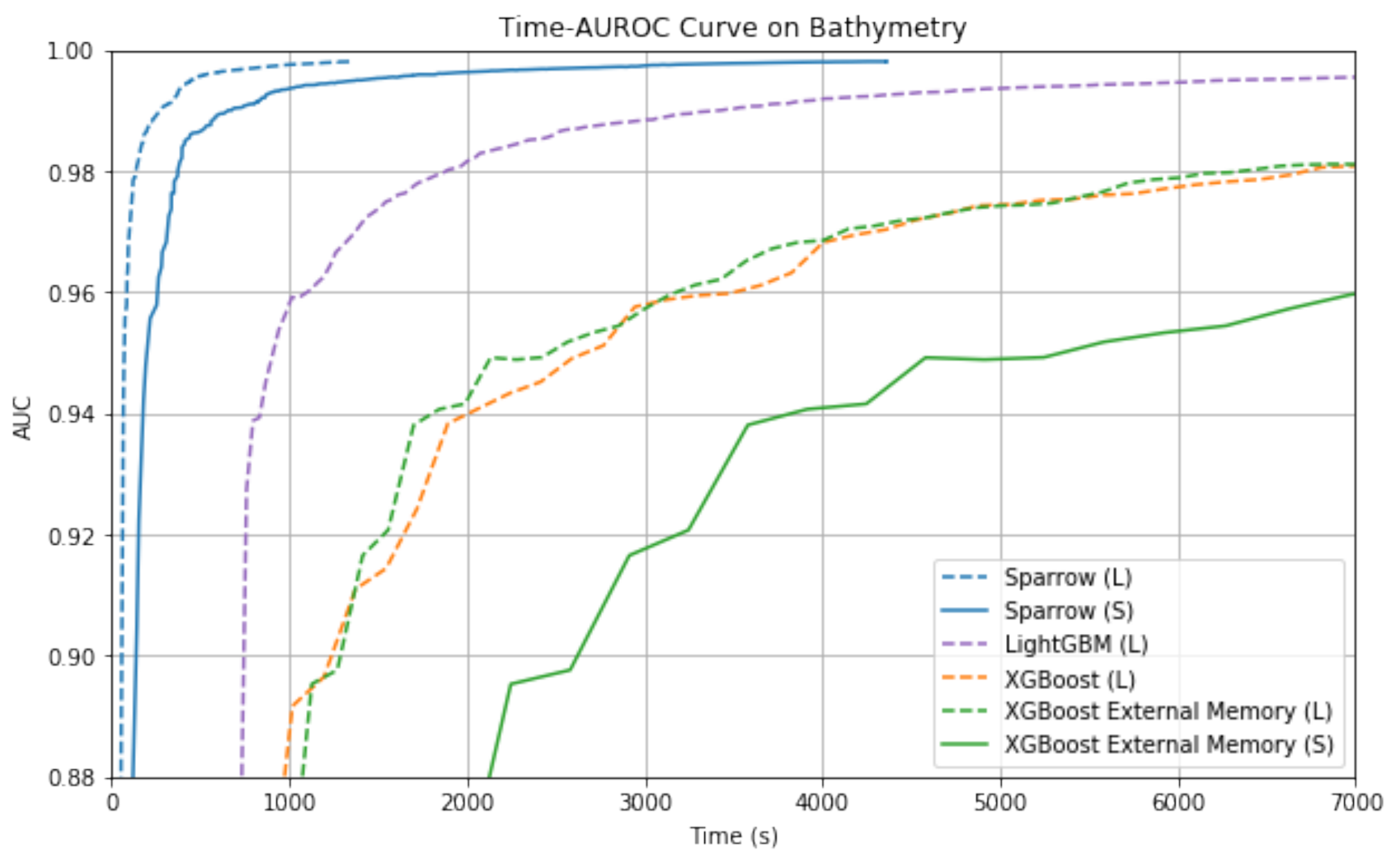}
    }
    \captionof{figure}{Time-AUROC curve on the bathymetry dataset, higher is better, clipped on right and bottom.
    The (S) suffix is for training on 61\,GB memory, and the (L) suffix is for training on 244\,GB memory.}~\label{fig:bathymetry-auroc}
\end{minipage}
\end{figure}

Due to the space limit, we put the detailed summary of the experiment results
in Table~\ref{table-splice} and Table~\ref{table-bathymetry}
in the Appendix~\ref{appendix:experiments}.
We evaluated each algorithm
in terms of the AUROC as a function of training time on the testing dataset.
The results are given in Figure~\ref{fig:splice-auroc} and
Figure~\ref{fig:bathymetry-auroc}.

On the splice site dataset, \Sparrow\ is able
to run on the instances with as small as 8\,GB memory.
The external memory version of XGBoost
can execute with reasonable amount of
memory (but still needs to be no smaller than 15\,GB)
but takes about 3x longer training time.
However, we also noticed that \Sparrow\ does not
have an advantage over other two boosting implementations
when the memory size is large enough to load in the entire
training dataset.

On the bathymetry dataset, \Sparrow\ consistently
out-performs XGBoost and LightGBM,
even when the memory size is larger than the dataset
size. In extreme cases, we see that
\Sparrow\ takes 10x-20x shorter training time and
achieves better accuracy.
In addition, both LightGBM and the in-memory version of
XGBoost crash when trained with less than 244\,GB memory.

We observed that properly initializing 
 the value of $\gamma$ and setting a reasonable
  sample set size can have great impact on the
  performance of \Sparrow.
  If stopping rule frequently fails to fire,
  it can introduce a significant overhead to the
  training process.
 Specific to the boosted trees,
 one heuristic we find useful is to initialize
  $\gamma$ to the maximum advantage of the tree nodes
  in the previous tree.
  A more systematic approach for deciding $\gamma$
  and sample set size is left as future work.

\section{Conclusion and Future Work}\label{sec:Conclusion}
In this paper, we have proposed a boosting algorithm contains
three techniques: {\em effective number of examples}, {\em weighted
sampling}, and {\em early stopping}.
Our preliminary results show that they
can dramatically speed up boosting algorithms on large
real-world datasets, especially when the data size
exceeds the memory capacity.
For future work,
we are working on a parallelized version of \Sparrow\ which
uses a novel type of asynchronous communication protocol.
It uses stopping rule to do model update, and
relaxes the necessity for frequent communication between
multiple workers especially when training on large datasets,
which we believe is a better parallel learning paradigm.

\newpage

\bibliography{paper}
\bibliographystyle{plain}

\newpage
\appendix

\setcounter{page}{1}

\section{Evaluate Sparrow on Large Datasets}\label{appendix:experiments}

Due to the space limit, we summarize the detailed training time in each experiment
in the appendix.

The experiments on large datasets are all conducted on EC2 instances with attached SSD storages from Amazon Web Services.
We ran the evaluations on five different instance types with increasing memory capacities, specifically
8\,GB (\texttt{c5d.xlarge}, costs \$0.192 hourly), 15.25\,GB (\texttt{i3.large}, costs \$0.156 hourly),
30.5\,GB (\texttt{i3.xlarge}, costs \$0.312 hourly), 61\,GB (\texttt{i3.2xlarge}, costs \$0.624 hourly), and 244\,GB (\texttt{i3.8xlarge}, costs \$2.496 hourly).

In Table~\ref{table-splice} and Table~\ref{table-bathymetry},
we compared the training time it takes to reduce the exponential loss as evaluated on the testing data.
Specifically, we compared the values of the average loss when the training converges and
the corresponding training time.
In addition,
we observed that
the average losses converge to slightly different values,
because two of the algorithms in comparison, Sparrow and LightGBM, apply sampling methods
during the training.
Therefore,
we also compared the training time it takes for each algorithm to reach the same threshold
for the average loss.

We use ``XGB'' for XGBoost, and ``LGM'' for LightGBM in the tables.
In addition, we observe that the training speed on the 8\,GB instances is better than that on 15\,GB instances,
because the 8\,GB instance has more CPU cores than the 15\,GB instance.

\begin{table}[]
\begin{minipage}[b]{.48\textwidth}
\centering
\captionof{table}{Training time (hours) on
the splice site dataset.
The (m) suffix is trained in memory.
The (d) suffix is trained with disk as external memory.
}\label{table-splice}
\vspace{0.5em}

{Training time until the loss convergences}
\begin{tabular}{|r|r|r|r|r|}
\hline
 Memory  & Sparrow   & XGB & LGM   \\  \hline
8\,GB & \textbf{2.9} (d)  & OOM             & OOM             \\ 
15\,GB & \textbf{8.4} (d)  & $>50$ (d)             & OOM             \\
30\,GB & \textbf{10.4} (d)   & 0.6  (d)   & OOM             \\
61\,GB & 4.4 (d)   & 12.8 (d) & \textbf{1.2} (m) \\
244\,GB & 1.3 (d)  & 1.1  (m) & \textbf{0.5} (m)  \\ \hline
Converged & 0.057 & 0.055 & \textbf{0.053} \\ \hline
\end{tabular}

\vspace{1em} 

{\small Training time until the average loss reaches $0.06$}
\begin{tabular}{|r|r|r|r|r|}
\hline
 Memory  & Sparrow   & XGB & LGM   \\  \hline
8\,GB & \textbf{1.4} (d)  & OOM       & OOM       \\
15\,GB & \textbf{7.1} (d) &  $>50$ (d)   & OOM       \\
30\,GB & \textbf{2.3} (d) & 9.3 (d) & OOM       \\
61\,GB &  1.3 (d) & 4.6 (d) & \textbf{0.3} (m) \\
244\,GB & 0.5 (d) & 0.3 (m) & \textbf{0.2} (m) \\ \hline
\end{tabular}

\end{minipage}
\hfill
\begin{minipage}[b]{.48\textwidth}
\centering
\captionof{table}{Training time (hours) on
the bathymetry dataset.
The (m) suffix is trained in memory.
The (d) suffix is trained with disk as external memory.}\label{table-bathymetry}
\vspace{0.5em}

{Training time until the loss convergences}
\begin{tabular}{|r|r|r|r|r|}
\hline
Memory & Sparrow   & XGB  & LGM    \\ \hline
8\,GB & \multicolumn{3}{c|}{The disk cannot fit the data} \\ \hline
15\,GB  &  \textbf{2.5} (d)  & OOM & OOM \\
30\,GB  &  \textbf{1.9}  (d)  & 41.7 (d) & OOM  \\
61\,GB  &  \textbf{1.2} (d)  & 38.6 (d) & OOM \\
244\,GB  &  \textbf{0.4} (d) & 20.0 (m) &  4.0 (m) \\  \hline	
Converged  & \textbf{0.046}        & 0.054       & 0.054 \\ \hline	
\end{tabular}

\vspace{1em}

{\small Training time until the average loss reaches $0.06$}
\begin{tabular}{|r|r|r|r|r|}
\hline
Memory & Sparrow   & XGB  & LGM    \\ \hline
8\,GB & \multicolumn{3}{c|}{The disk cannot fit the data} \\ \hline
15\,GB   & \textbf{1.0} (d)  & OOM       & OOM         \\
30\,GB   & \textbf{0.6} (d)  & 41.7 (d)        & OOM         \\
61\,GB   & \textbf{0.6} (d)  & 38.4 (d)        & OOM         \\ 
244\,GB   & \textbf{0.2} (d)  & 16.9 (m)  & 3.3 (m) \\  \hline
\end{tabular}

\end{minipage}
\end{table}

\section{Stopping rule analysis} \label{sec:balsubramani}

We set the stopping rule applied in Sparrow (Equation~\ref{eqn:stopping_rule}) based on the following theorem.

\begin{theorem}[based on Balsubramani~\cite{balsubramani_sharp_2014} Theorem 4] \label{thm:balsubramani}
  Let $M_t$ be a martingale $M_t = \sum_i^t X_i$,
  and suppose there are constants $\{c_k\}_{k \geq 1}$ such that
  for all $i \geq 1$, $|X_i| \leq c_i$ w.p.\ 1.
  For $\forall \sigma > 0$, with probability at least $1 - \sigma$ we have
  \[
  \forall t: |M_t| \leq C \sqrt{
    \left( \sum_{i=1}^t c_i^2 \right)
    \left( \log \log \left( \frac{ \sum_{i=1}^t c_i^2 }{ |M_t| }\right) +
    \log \frac{1}{\sigma} \right)
  },
  \]
  where $C$ is a universal constant.
\end{theorem}

In our experiments, we set $C = 1$ and $\sigma=\frac{0.001}{{|\cH|}}$,
where $\cH$ is the set of base classifiers (weak rules).

\section{Pseudocode for \Sparrow}\label{appendix:pseudocode}


\begin{algorithm}[H]
\caption{Main Procedure of \Sparrow}\label{algorithm}

\begin{algorithmic}[0]
\STATE {\textbf{Input} Sample size $\n$}
\STATE {\hspace{2.5em} A threshold $\theta$ for the minimum $\neff /\n$ ratio for training weak learner}
\STATE
\STATE \textbf{Initialize} $H_{0} =0$
\STATE \textbf{Create} initial sample $S$ by calling \textsc{Sample}
\FOR{$k:=1 \ldots K$}
  \STATE Call \textbf{Scanner} on sample $S$ generate weak rule $h_k,\gamma_k$
  \STATE $H_k \gets H_{k-1} + \frac{1}{2} \log \frac{1/2+\gamma}{1/2-\gamma} h_k$
  \IF{$\neff /\n<\theta$}
  \STATE {Receive a new sample $S \gets$ from \textbf{Sampler}}
  \STATE {\textbf{Set} $S \gets S'$ }
  \ENDIF
\ENDFOR

\end{algorithmic}

\end{algorithm}

\begin{algorithm}[H]

\caption{Scanner}\label{alg-scanner}

\begin{algorithmic}[0]
\STATE {\textbf{Input} An iterator over in-memory sampled set $S$}
\STATE {\hspace{2.5em} Initial advantage target $\gamma_0 \in (0.0, 0.5)$}
\STATE

\STATE {\textbf{static variable} $\gamma = \gamma_0$}

\LOOP

\IF{sample $S$ is scanned without firing stopping rule}
   \STATE Shrink $\gamma$ by $\gamma \gets 0.9\,\widehat{\gamma}$
   \STATE Reset $S$ to scan from the beginning
\ENDIF

\STATE $(x, y, w_l) \gets S.next()$
\STATE $w \gets \textsc{UpdateWeight}(x, y, w_l, H)$
   
\FOR{$h \in \weakRules$}
\STATE Compute $h(\vx) y$
\STATE Update $M_t,V_t$ (Eqn~\ref{eqn:stopping-rule-statistics})

\IF{Stopping Rule (Eqn~\ref{eqn:stopping_rule}) fires}
   \STATE \textbf{return} $h,\gamma$
\ENDIF
\ENDFOR

\ENDLOOP


\end{algorithmic}

\end{algorithm}

\begin{algorithm}[H]

\caption{Sampler}\label{alg-sampler}

\begin{algorithmic}[0]

\STATE \textbf{Input } Randomly permuted, disk-resident training-set \\
\STATE

\STATE {Disk-resident stratified structure} $D \gets \{\}$
\STATE {Weights of the strata} $W \gets \{\}$
\STATE Construct new sample $S \gets \{\}$

\LOOP
\STATE With the probability proportional to $W$,
\STATE \hspace{1cm} \textbf{sample} a strata $R$
\STATE $(x, y, w_l) \gets R.next()$
\STATE \textbf{Delete} $(x, y, w_l)$ from $R$, \textbf{update} $W$

\STATE Receive new model $H$ from \textsc{MainProcedure}
\STATE $w \gets \textsc{UpdateWeight}(x, y, w_l, H)$

\STATE \textrm{With the probability proportional to } $w$, \\
	\hspace{1cm} $S \gets S + \{( x, y, w )\}$.

\STATE \textbf{Append} $(x, y)$ to the right stratum with regard to $w$, \\
\hspace{1cm} $D \gets D + \{( x, y, w ) \}$ \\
\hspace{1cm} \textbf{Update} $W$

\IF { $S$ is full }
\STATE \textbf{Send } $S$ to \textsc{MainProcedure}
\STATE $S \gets \{\}$
\ENDIF

\ENDLOOP


\end{algorithmic}

\end{algorithm}






\end{document}